\documentclass{article}
\usepackage{ijcai24}

\usepackage{times}
\usepackage{soul}
\usepackage{url}
\usepackage[hidelinks]{hyperref}
\usepackage[utf8]{inputenc}
\usepackage[small]{caption}
\usepackage{graphicx}
\usepackage{amsmath}
\usepackage{amsthm}
\usepackage{booktabs}
\usepackage{algorithm}
\usepackage{algorithmic}
\usepackage[switch]{lineno}

\newtheorem{definition}{Definition}

\usepackage{esvect}
\usepackage{listings}
\usepackage{float}
\usepackage{multirow}
\usepackage{xcolor}
\usepackage{wrapfig}

\usepackage{xspace}

\newcommand{\ie}{\textit{i.e.}\xspace}

\newcommand{\eg}{\textit{e.g.}\xspace}
\newcommand{\etc}{\textit{etc.}\xspace}

\newenvironment{sloppypar*}
 {\sloppy\ignorespaces}
 {\par}

\lstdefinestyle{mypystyle}{
    language=Python,
    basicstyle=\small\ttfamily,
    keywordstyle=\color{blue},
    stringstyle=\color{red},
    commentstyle=\color{green},
    breaklines=true,
    showstringspaces=false,
    morekeywords={True, False},
    frame=single,
}

\title{Contextual Importance and Utility in Python: \\New Functionality and Insights with the \texttt{py-ciu} Package}

\author{
    Kary Fr\"amling
    \affiliations
    Umeå University, Sweden; Aalto University, Finland
    \emails
    kary.framling@cs.umu.se
}

\begin{document}

\maketitle

\begin{abstract}
The availability of easy-to-use and reliable software implementations is important for allowing researchers in academia and industry to test, assess and take into use eXplainable AI (XAI) methods. This paper describes the \texttt{py-ciu} Python implementation of the Contextual Importance and Utility (CIU) model-agnostic, post-hoc explanation method and illustrates capabilities of CIU that go beyond the current state-of-the-art that could be useful for XAI practitioners in general. 

\end{abstract}

\section{Introduction}

Explainable AI (XAI) has surged in significance in recent years as AI systems permeate various facets of our lives, notably through novel achievements of Machine Learning (ML) methods. Understanding why AI makes certain decisions is crucial, particularly in fields like healthcare, finance, and criminal justice, where transparency and accountability are paramount for maintaining the trust in such AI systems. 

Feature attribution methods such as the Local Interpretable Model-agnostic Explanations (LIME) \cite{ribeiro2016_LIME} and SHapley Additive exPlanations (SHAP) \cite{NIPS2017_Lundberg_XAI} seem to be the most popular XAI methods for the moment, at least in the category of so called model-agnostic post-hoc (/outcome) explanation methods. The popularity of those methods might partially be due to the availability of open source implementations, which makes it possible to use them relatively easily in research and in real-life applications. 

In this paper we present a Python implementation for tabular data of the Contextual Importance and Utility (CIU) method. Like LIME and SHAP, CIU is a model-agnostic post-hoc explanation method. However, CIU's theoretical foundation is different from both LIME, SHAP and most current XAI methods by making a difference between \textit{feature importance} and \textit{feature influence}, where feature influence conceptually corresponds to the values produced by LIME and SHAP. This theoretical difference makes it possible to produce different explanations than other XAI methods, such as the \textit{potential influence plot} presented in section~\ref{Sec:PotentialInfluencePlot}. 

The objectives of the paper are twofold:

\begin{enumerate}
    \item Introduce a Python implementation of CIU for tabular data that is comparable with those that exist for LIME and SHAP.
    \item Using the presented implementation, demonstrate in what way CIU differs from other methods and what additional capabilities it offers in terms of explainability.  
\end{enumerate}

After this Introduction, Section~\ref{Sec:CIU} resumes the core theory of CIU. Section~\ref{Sec:SotA} shows how \texttt{py-ciu} can produce similar explanations as state-of-the-art methods. Section~\ref{Sec:NewFunctionality} shows functionality that is specific to CIU, followed by Conclusions. 

\section{Contextual Importance and Utility}\label{Sec:CIU}

\begin{figure}
    \centering
    \includegraphics[width=\columnwidth]{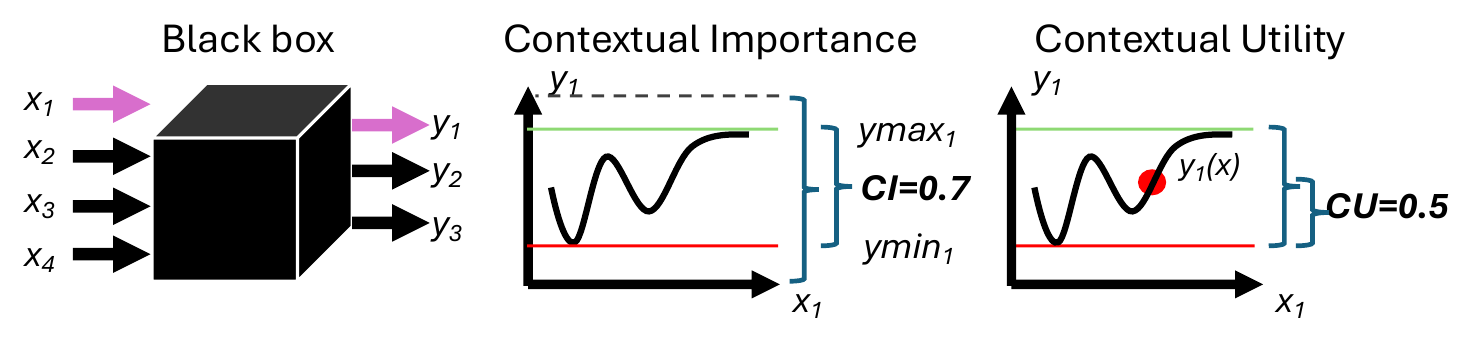}
    \caption{Illustration of how CI and CU values are calculated for the input-output pair $x_{1}$ and $y_{1}$.}
    \label{fig:figciu}
\end{figure}

Contextual Importance and Utility (CIU) was initially presented by Kary Fr\"{a}mling in 1992 \cite{FramlingDEA_1992} for explaining recommendations of decision support systems in a model-agnostic way, where ``model-agnostic'' includes non-ML based models. CIU was presented formally in \cite{FramlingAISB_1996,FramlingThesis_1996}. More recent research on CIU has been presented \eg in \cite{AnjomshoaeEtAl_CIU_2019,FramlingAJCAI2022,Framling_xAI_2023}. 
Figure~\ref{fig:figciu} illustrates how the two core concepts Contextual Importance (CI) and Contextual Utility (CU) are calculated for the input feature $x_{1}$ and the output $y_{1}$. The only values that need to be calculated or estimated are $ymin_{j}$ and $ymax_{j}$. The plots in Figure~\ref{fig:Titanic_IO_plots} are generated by \texttt{py-ciu} and illustrate how CI and CU values have been calculated in the same way as Figure~\ref{fig:figciu}.

CIU uses the notion of \textit{utility} known from domains such as utility theory, multiple criteria decision making \etc \cite{KeeneyRaiffa_1976,roy1book85,Vin92,koksalan2011multiple}. Utility theory is postulated in economics to explain behavior of individuals based on the premise that people can consistently rank order their choices depending upon their preferences. Each individual will show different preferences. Under some assumptions, those preferences can be represented analytically using a \textit{utility function}. For the purposes of this paper, it is adequate to say that a utility function maps values $x_{\{i\}}$ or $y_{j}$ into a utility value $u\in[0,1]$. For instance, the  utility value of output $y_{j}$ is $u_{j} = h_{j}(y_{j})$, where $h_{j}$ is the utility function for output $y_{j}$. In classification tasks, we have $u_{j} = y_{j}$ because $y_{j}$ is typically a probability that can be used directly as the utility value $u_{j}$.

We begin with the formal definitions of CI and CU and then go on with a definition of \textit{Contextual influence}, denoted with the symbol $\phi$.

\begin{definition} [Contextual Importance (CI)] CI expresses to what extent modifying the value of one or more feature(s) $x_{\{i\}}$ can modify the output utility value $u_{j}$. CI is expressed formally as:

\begin{eqnarray}
CI_{j}(x,\{i\},\{I\})=\frac{umax_{j}(x,\{i\})-umin_{j}(x,\{i\})}{umax_{j}(x,\{I\})-umin_{j}(x,\{I\})},  
\label{Eq:GeneralCI}
\end{eqnarray}

where $x$ is the studied instance, $\{i\} \subseteq \{I\}$ and $\{I\} \subseteq \{1,\dots,n\}$, and $n$ is the number of features. $umin_{j}$ and $umax_{j}$ are the minimal and maximal output utility values that can be achieved by varying the value(s) of feature(s) $x_{\{i\}}$ while keeping all other feature values at those of $x$.
\end{definition}

\noindent For classification tasks where $u_{j}(y_{j})=y_{j}\in[0,1]$ and for regression tasks where the utility value can be obtained by an affine transformation of form $u_{j}(y_{j})=Ay_{j}+b$ (which applies to most regression tasks), we can write: 

\begin{eqnarray}
CI_{j}(x,\{i\},\{I\})= \frac{ymax_{j}(x,\{i\})-ymin_{j}(x,\{i\})}{ ymax_{j}(x,\{I\})-ymin_{j}(x,\{I\})}, 
\label{Eq:CI}
\end{eqnarray}

where $ymin_{j}$ and $ymax_{j}$ are the minimal and maximal output values observed when varying the value(s) of feature(s) $x_{\{i\}}$ and $x_{\{I\}}$. When $\{I\}=\{1,\dots,n\}$, \ie all features, then $ymax_{j}(x,\{I\})$ and $ymin_{j}(x,\{I\}$ rarely need to be estimated because they are one and zero for classification tasks. For regression tasks, they can be set to the minimal and maximal output values in the training set. 

\begin{definition} [Contextual Utility (CU)] CU expresses how the current value(s) of feature(s) $x_{\{i\}}$  contribute to obtaining a high output utility $u_{j}$: 

\begin{eqnarray}
CU_{j}(x,\{i\})=\frac{u_{j}(c)-umin_{j}(x,\{i\})}{umax_{j}(x,\{i\})-umin_{j}(x,\{i\})} 
\label{Eq:CU}
\end{eqnarray}
\end{definition}

\noindent When $u_{j}(y_{j})=Ay_{j}+b$, this can again be written as:  

\begin{eqnarray}
CU_{j}(x,\{i\})=\left|\frac{ y_{j}(c)-yumin_{j}(x,\{i\})}{ymax_{j}(x,\{i\})-ymin_{j}(x,\{i\})}\right|, 
\label{Eq:CU_y}
\end{eqnarray}

where $yumin=ymin$ if $A$ is positive and $yumin=ymax$ if $A$ is negative. 

\begin{definition} [Contextual influence] Contextual influence expresses how much one or more feature(s) influence the output value (utility) relative to a \textit{reference value} or \textit{baseline}, here denoted $CU_{ref} \in [0,1]$. Contextual influence is defined as: 
\begin{equation}
\phi=CI \times (CU - CU_{ref}),
\label{Eq:Contextual_influence}
\end{equation}

where indices ``$_{j}(x,\{i\},\{I\})$'' have been omitted for easier readability. 
\end{definition}

Contextual influence is conceptually similar to Shapley value and similar additive feature attribution methods, which is a reason to use the symbol $\phi$. 

CIU equations apply to individual features as well as to coalitions of features $\{i\}$ versus other coalitions of features $\{I\}$, where $\{i\} \subseteq \{I\}$ and $\{I\} \subseteq \{1,\dots,n\}$. When such coalitions are given labels, they can be used as \textit{Intermediate Concepts (IC)} in the explanations. Such ICs make it possible to define arbitrary explanation vocabularies with abstraction levels that can be adapted to the target user. 

It is worth noting that CI and CU are values in the range $[0,1]$ by definition. Such a known value range makes it possible to assess whether a value is high or low. Contextual influence also has a maximal amplitude of one, where the range is $[-CU_{ref}, 1-CU_{ref}]$. 


\subsection{Estimation of \texorpdfstring{$ymin$}{Lg} and \texorpdfstring{$ymax$}{Lg}}

CIU only depends on identifying sufficiently correct $ymin$ and $ymax$ values for the studied model, instance and set of features $\{i\}$, and $\{I\}$ if needed. In \texttt{py-ciu} the class \texttt{PerturbationMinMaxEstimator} takes care of searching for $ymin$ and $ymax$ values in a model-agnostic way. \texttt{PerturbationMinMaxEstimator} generates a set of samples for which $y$ values are calculated. 
The principle of building the set of samples is as follows: 

\begin{enumerate}
    \item For categorical features in $\{i\}$, include all possible value combinations. 
    \item For numerical features in $\{i\}$, include all possible combinations of minimal and maximal input values.
    \item Combine all the possible value combinations for categorical and numerical features from the two previous steps.
    \item If the number of samples so far is smaller than the number requested, then fill up with random values for numerical features in $\{i\}$, and for random combinations of categorical features values in $\{i\}$.
    \item Concatenate current instance with the other samples. 
\end{enumerate}

\noindent \texttt{PerturbationMinMaxEstimator} is the default class for estimating $ymin$ and $ymax$. The default number of samples is 100, which in most cases is a good compromise when dealing with only one feature $x_{i}$. However, identifying minimal and maximal function values could be done in many different ways. It would, for instance, be possible to use an estimator that only produces in-distribution samples. Information about the model $f$ could also be used if available. If $f$ is linear, then only two samples are needed per feature. For rule- or tree-based systems it would presumably be possible to identify $ymin$ and $ymax$ efficiently by known information about their properties, as for TreeSHAP \cite{LundbergEtAl_TreeSHAP_2018}. In \cite{FramlingAISB_1996}, a Radial Basis Function (RBF) neural network architecture was used, where $ymin$ and $ymax$ values were by definition found at, or close to, the RBF centroids. The estimation of $ymin$ and $ymax$ is a mathematical task that is not specific to CIU and where significant improvements can be expected in future research.

\subsection{Out Of Distribution?}

\begin{figure*}[!ht]
    \centering
    \includegraphics[width=0.33\textwidth]{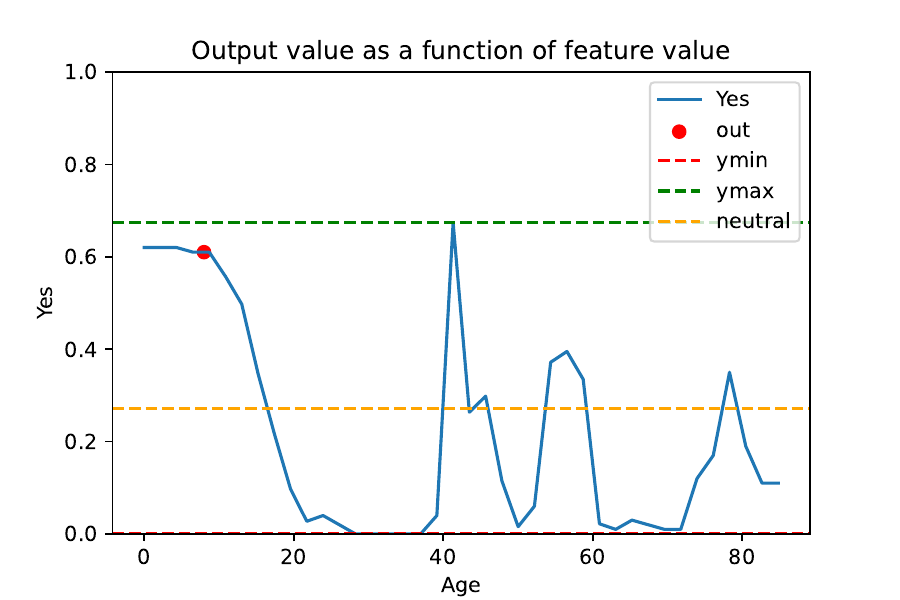}
    \includegraphics[width=0.33\textwidth]{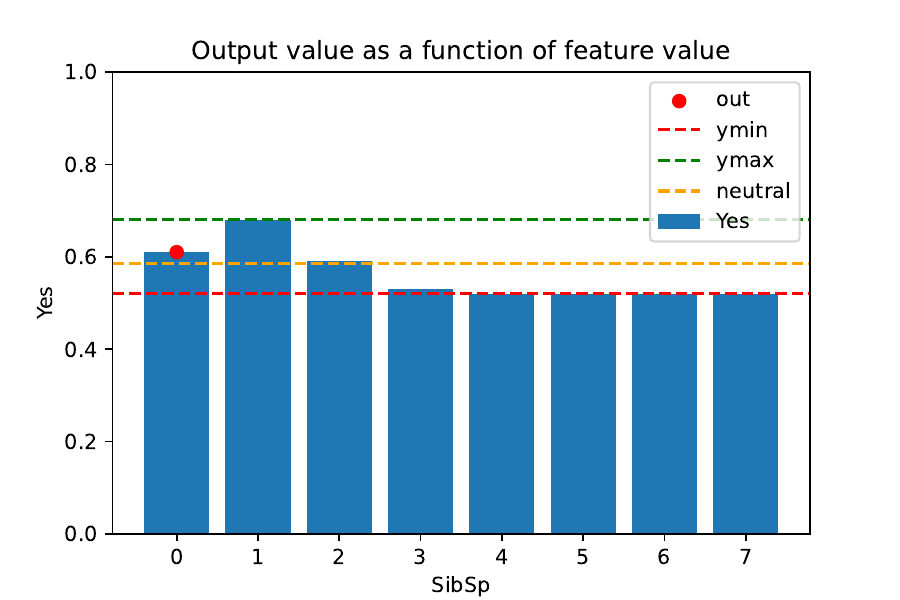}
    \includegraphics[width=0.33\textwidth]{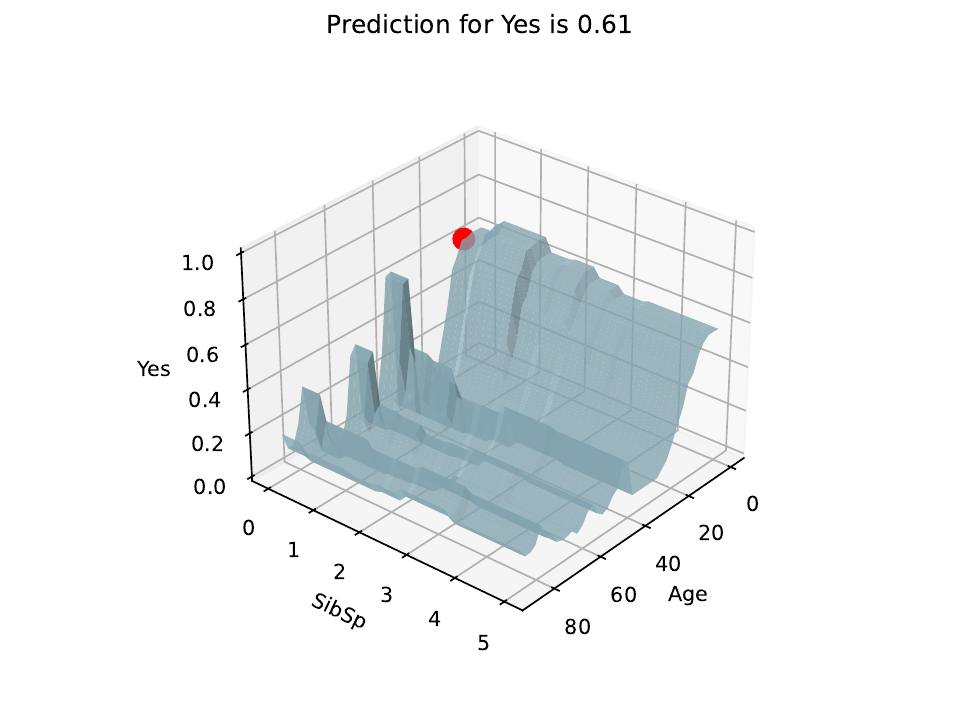}
    \caption{Input-Output plots showing the probability of survival as a function of the numeric feature Age, the categorical feature Sibsp (number of siblings) and 3D plot of them jointly. CIU illustration has been included in the 2D plots by setting the parameter \texttt{illustrate\_CIU=True}.}
    \label{fig:Titanic_IO_plots}
\end{figure*}

The sampling approach described in the previous section can generate so called out of distribution (OOD) samples, \ie feature value combinations that are not possible in reality or that are significantly different from the data in the training set used to build the ML model being explained. For such samples, the model $f$ may be incapable to provide correct output values. OOD challenges related to the used sampling method and potential solutions to those challenges can be grouped into the three following cases:

\begin{enumerate}
    \item \textbf{Predictable OOD behaviour.}  If OOD samples do not lead to undershooting of the $ymin$ value, nor to overshooting of the $ymax$ value, then OOD is not an issue. Many ML models do not under- or overshoot even when extrapolating outside the training set, \eg Linear Discriminant Analysis (LDA), and most rule- or tree-based methods such as random forest and gradient boosting. Neural network models such as the Interpolating, Normalising and Kernel Allocating (INKA) used in \cite{FramlingAISB_1996} also guarantee that under- or overshooting does not occur. Input-output value graphs such as those in Figure~\ref{fig:Titanic_IO_plots} can be used for studying the model behaviour within the value ranges used by CIU. 
    \item \textbf{Non-predictable OOD behaviour.} This happens if under- or overshooting occur with OOD samples. In that case the sampling approach used here will not be appropriate. Various approaches could be imagined for addressing this problem, such as only using samples that are ``sufficiently'' close to instances in the training set. 
    \item \textbf{Detecting model instability}. Since CI and CU values are in the range $[0,1]$ by definition, obtaining CI or CU values that are outside this range indicate that the model undershoots or overshoots the permissible range for one or more samples. This could be an indication that those samples should be removed or that the model should be corrected in order to increase its trustworthiness. A correction approach using so called pseudo-examples has been proposed \eg in \cite{FramlingThesis_1996}. 
\end{enumerate}

\noindent The second and third cases are left out of scope for the current paper and remain topics of further research. Similar OOD challenges exist for all permutation-based XAI methods, including kernel-SHAP and LIME. 


\section{State of the Art}\label{Sec:SotA}

\begin{figure*}
    \centering
    \includegraphics[width=0.32\textwidth]{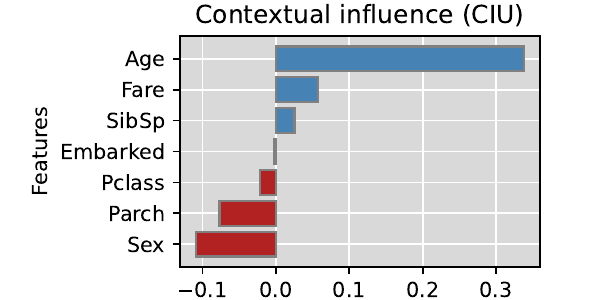}
    \includegraphics[width=0.32\textwidth]{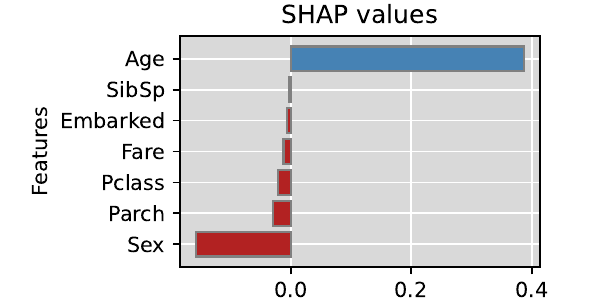}
    \includegraphics[width=0.32\textwidth]{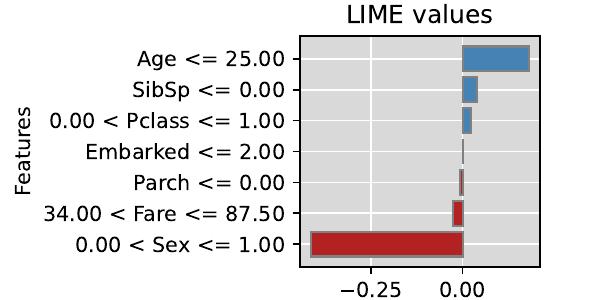}
    \caption{Influence explanations with CIU (contextual influence), SHAP and LIME for Johnny D's probability of survival (61\%).}
    \label{fig:Titanic_influence_plots}
\end{figure*}

CIU was originally implemented in Matlab as described in \cite{FramlingThesis_1996} and \cite{AnjomshoaeEtAl_CIU_2019}. Since then, CIU has been implemented in \texttt{R} for tabular data \cite{Framling_XAI_WS_AAAI_2021} and image classification \cite{Framling_ciuimageR_EXTRAAMAS_2021}\footnote{The \texttt{R} implementation has later been superseded by a Python version \cite{Framling_py_ciu_image_EXTRAAMAS_2024}}. A first Python implementation of CIU for tabular data was published and described in \cite{AnjomshoaeEtAl_PyCIU}. However, that implementation was very ``bare-bone'' and instantly lagged behind the functionality available in the \texttt{R} version. 

This paper describes the functionality of \texttt{py-ciu} after a complete rewrite performed in the end of 2023. The goal of the rewrite was to take the Python implementation of CIU for tabular data to at least the same level as the corresponding \texttt{R} implementation, as it is described in \cite{Framling_CIU_EXTRAAMAS_2023}. The intention has also been to make the use of the \texttt{R} and Python implementations as similar as possible. In the remainder of this section, we show how to produce well-known visualisations and explanations with \texttt{py-ciu}. \texttt{py-ciu} is available at \url{https://github.com/KaryFramling/py-ciu} and the code that has been used for producing the results of this paper are in the notebook \texttt{XAI\_IJCAI\_2024.ipynb} of the repository.

To begin with, we will use the well-known Titanic data set and a Random Forest classifier model for predicting the probability of survival. The explained instance is ``Johnny D'', who is also used in \url{https://ema.drwhy.ai}. ``Johnny D'' is an 8-year old boy who travels alone. The model predicts a 61.0\% probability of survival, which is significantly higher than the average probability of survival 40.4\% for the data set used. We want to explain why the estimated probability of survival is rather high.

Listing~\ref{lst:CIUinit} shows how to initialise a CIU object and how to get a CIU result. The result \texttt{CIUres\_titanic} contains all the information needed for producing plots, textual explanations or other kinds of explanations. The documentation of all classes, methods, parameters \etc are available in the Sphinx-generated pages  at \url{https://py-ciu.readthedocs.io/}.  


\begin{lstlisting}[style=mypystyle, caption={Creation of CIU object and getting the CIU result as a pandas DataFrame.}, label=lst:CIUinit]
CIU_titanic = ciu.CIU(titanic_model.predict_proba, ['No', 'Yes'], data=titanic_train, category_mapping=category_mapping, neutralCU=mean_surv_prob)
CIUres_titanic = CIU_titanic.explain(new_passenger, output_inds=1)
\end{lstlisting}

In Listing~\ref{lst:CIUinit}, the parameters that are passed to the constructor are: 
\begin{itemize}
    \item \texttt{titanic\_model.predict\_proba}: The model's prediction function to use. For classification models, this function should return class probabilities. 
    \item \texttt{['No', 'Yes']}: Names of the output classes. 
    \item \texttt{data}: Training data as a pandas DataFrame. CIU currently only uses this parameter for determining the value ranges of input features. Another alternative is to pass those ranges as a parameter. 
    \item \texttt{category\_mapping}: Dictionary that contains names and possible values of features that should be dealt with as categories, i.e. having discrete int/str values. 
    \item \texttt{neutralCU}: $CU_{ref}$ value for contextual influence. 
\end{itemize}

The parameters to the \texttt{explain()} method are the instance to explain (a pandas DataFrame) and the index of the model output to explain. \texttt{explain()} returns a pandas DataFrame with the CIU results. The last CIU result is retained and used by default, so it does not have to be passed as a parameter to the explanation-generating methods in the next sections. 

\subsection{Input-Output Plots}

Plotting the output value $y_{j}$ as a function of one (or two) inputs $x_{\{i\}}$ while keeping the values of all other inputs static is a rather obvious way of studying the behaviour of any model. Such \textit{input-output (IO) plots} are used extensively in CIU papers, such as \cite{FramlingThesis_1996,FramlingAISB_1996}. Later, \cite{Friedman_PDP_2001} suggested to use the name \textit{Partial Dependence Plot (PDP)} for such visualisations. Names such as \textit{Individual Conditional Expectation (ICE)} plots and \textit{Ceteris Paribus (PB)} plots have also been suggested for essentially the same thing. In this paper, we will simply call them IO plots because that name seems to be oldest in a XAI context and is also the name associated with CIU. Figure~\ref{fig:Titanic_IO_plots} shows IO plots generated by the commands shown in Listing~\ref{lst:IOplots}. 

\begin{lstlisting}[style=mypystyle, caption={Generating IO plots.}, label=lst:IOplots]
CIU_titanic.plot_input_output(ind_input=2, output_inds=1, illustrate_CIU=True)
CIU_titanic.plot_input_output(ind_input=3, output_inds=1, illustrate_CIU=True)
CIU_titanic.plot_3D([2,3], ind_output=1, azim=40)
\end{lstlisting}

\noindent As illustrated in Figure~\ref{fig:figciu} and Figure~\ref{fig:Titanic_IO_plots}, CI, CU and contextual influence values can be ``read'' directly from IO plots: CI is the ratio $(ymax-ymin)/(whole\_y\_range)$, while CU corresponds to the position of the red dot (current instance values) within the interval $[ymin, ymax]$. Contextual influence corresponds to the y-axis position of the red dot relative to the orange ``neutral'' line that corresponds to the value of $CU_{ref}$, while being constrained by the red and green $ymin$ and $ymax$ lines. In the case of several outputs, which then typically represent different classes, \texttt{py-ciu} allows for including more than one or even all outputs $y_{j}$ in IO plots, which is useful for detecting where the transition happens between different classes. 

The 3D plot does not currently support visualising $ymin$, $ymax$ and $CU_{ref}$, mainly because introducing multiple planes in the 3D plot might make it more challenging to read the plot. 3D plots are useful because they make it possible to show dependencies between two features at a time. For instance, in Figure~\ref{fig:Titanic_IO_plots} we can see that it would be better for Johnny D to have one sibling rather than zero, whereas having no siblings seems to be favorable for survival for most other ages according to the model.

\subsection{Feature Influence Explanations}

Figure~\ref{fig:Titanic_influence_plots} shows the potentially most used kind of XAI plot for tabular data. The leftmost plot uses CIU's contextual influence and has been produced with the command \texttt{CIU\_titanic.plot\_influence()}. It is worth noting that in Listing~\ref{lst:CIUinit}, we gave the parameter \texttt{neutralCU=mean\_surv\_prob} for setting the value of $CU_{ref}$ to the average output probability rather than using the default value $CU_{ref}=0.5$. This was done in order to use the same reference value for contextual influence as SHAP does. 

Comparing these influence-based explanations given by different methods is out of the scope of this paper. However, it is worth noting that contextual influence has a known range of one between the lowest and highest possible $\phi$ values. That signifies that it is possible for the explainee to assess ``how negative'' or ``how positive'' a contextual influence value is even without seeing the $\phi$ values of other features. On the other hand, SHAP and LIME values do not have a pre-defined interpretation of $\phi$'s magnitude and mainly give an indication of relative influence between features. 





\section{CIU-specific Capabilities} \label{Sec:NewFunctionality}

By ``CIU-specific'' we mean XAI capabilites that are not possible or that have not been proposed with LIME, SHAP or other methods, to our best knowledge. 

\subsection{Potential Influence Plot}\label{Sec:PotentialInfluencePlot}

Explanations based on influence values as in Figure~\ref{fig:Titanic_influence_plots} can be misleading especially for instances that are similar or close to the reference instance or reference value. As an example, for an average instance with average feature values, all SHAP values might be zero, which could be called a \textit{null explanation}. The same is true also for Contextual influence and LIME values. This signifies that even the most important feature according to CI might have a zero influence value, which might lead to  misunderstandings with the explainee. 

\begin{wrapfigure}{r}{0.6\columnwidth}
\centering \includegraphics[width=0.6\columnwidth]{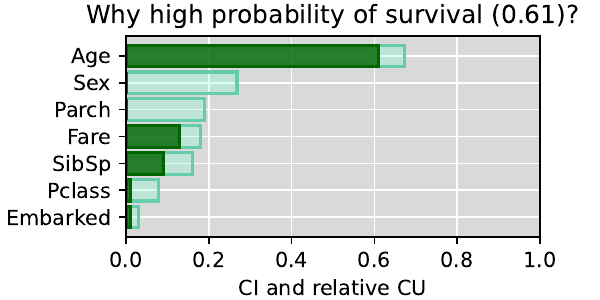}
    \caption{PI plot for 8-year old boy's probability of survival (61\%). }
    \label{fig:TitanicCIU_overlap}
\end{wrapfigure}

The \textit{Potential Influence (PI) plot} avoids null explanations, as shown in Figure~\ref{fig:TitanicCIU_overlap}. 
A PI plot illustrates the CI value with a transparent bar and overlays is with a solid bar that covers the transparent bar with as many percent as indicated by the CU value. Since CI and CU have know value ranges $[0,1]$, we can interpret them directly: $CI=0$ signifies no importance and no transparent bar, whereas $CI=1$ gives a transparent bar that fills the range entirely. Similarly, $CU=0$, signifies the worst possible value and no solid bar, whereas $CU=1$ gives a solid bar that covers the transparent bar entirely. Figure~\ref{fig:TitanicCIU_overlap} was produced with the command \texttt{CIU\_titanic.plot\_ciu(plot\_mode='overlap')}.

\begin{wrapfigure}{r}{0.6\columnwidth}
    \centering \includegraphics[width=0.6\columnwidth]{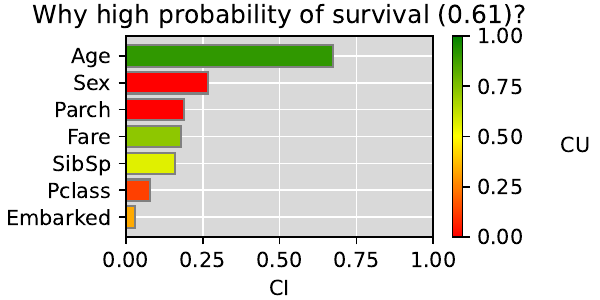}
    \caption{PI plot using bar length for visualizing CI and color for visualising CU, for 8-year old boy's probability of survival (61\%). }
    \label{fig:TitanicCIU}
\end{wrapfigure}

The reason for calling this a ``potential influence'' plot is that the area of the transparent part of each bar indicates the potential for improvement that can be achieved by modifying the value of that feature (or Intermediate Concept). The area of the solid part of each bar again indicates the potential for getting a worse result by modifying the value of the feature.

Figure~\ref{fig:TitanicCIU} shows an alternative PI visualisation using colours for indicating the CU value. Such a visualisation might be preferred in some contexts and by some users but we find that it doesn't indicate the potential influence of changing the feature value as clearly as Figure~\ref{fig:TitanicCIU_overlap}. 

In \cite{Framling_CIU_EXTRAAMAS_2023} the PI plot is described with the term ``counterfactual'' because it provides a partial answer to the question ``what if?''. The PI plot indeed gives an indication of which changes in feature values would have the greatest impact on the result. For instance, in Figure~\ref{fig:TitanicCIU_overlap} we see that having one sibling, rather than zero, could increase the probability of survival. However, knowing what exact changes to make, which changes are possible, or the cost of making them requires additional information. Furthermore, changing feature values will presumably also change CI and CU values due to the changing ``context'' as defined by $x$. Still, PI plots do provide guidance about what changes are worth trying first among the potentially numerous counterfactual options.

\subsection{Textual explanations}

\begin{figure*}[!ht]
    \centering \includegraphics[width=0.85\textwidth]{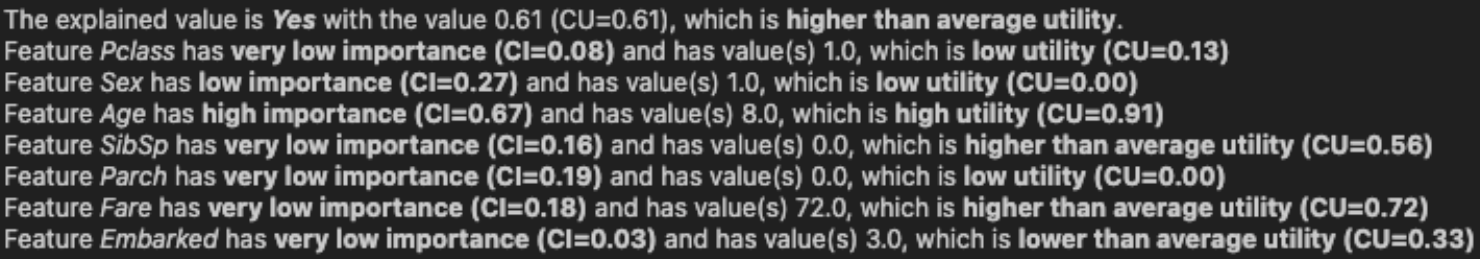}
    \caption{Textual CIU explanation using Markdown and text effects for Johnny D's probability of survival (61\%). }
    \label{fig:TitanicTextual}
\end{figure*}

Figure~\ref{fig:TitanicTextual} shows the output obtained from the commands in Listing~\ref{lst:TextualMardown}. The command \texttt{CIU\_titanic.textual\_explanation()} by itself produces raw text. Since CI and CU are constrained to the interval $[0,1]$, their values have a meaning that can be given a textual quantification. The thresholds and texts to use can be passed as parameters to the method. 

\begin{lstlisting}[style=mypystyle, caption={Producing a textual explanation with Markdown style.}, label=lst:TextualMardown]
markdown_text = CIU_titanic.textual_explanation(use_markdown_effects=True)
Markdown(markdown_text)
\end{lstlisting}

\subsection{Beeswarm Plot for Global Assessment}

\begin{wrapfigure}{r}{0.6\columnwidth}
    \centering
    \includegraphics[width=0.6\columnwidth]{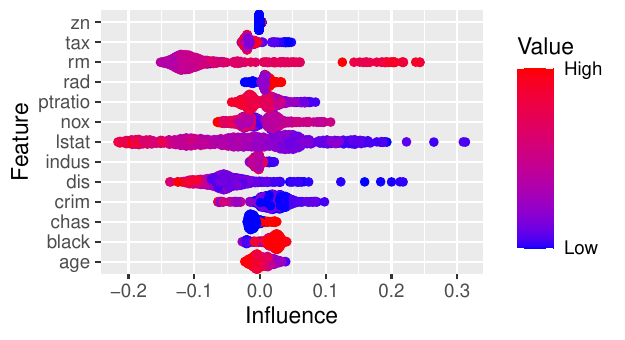} 
    \caption{Beeswarm plot from the \texttt{R} CIU package of Contextual influence for Boston data set.}
    \label{Fig:BeeswarmBostonR}
\end{wrapfigure}

\begin{figure*}[htb]
    \centering 
    \includegraphics[width=0.31\textwidth]{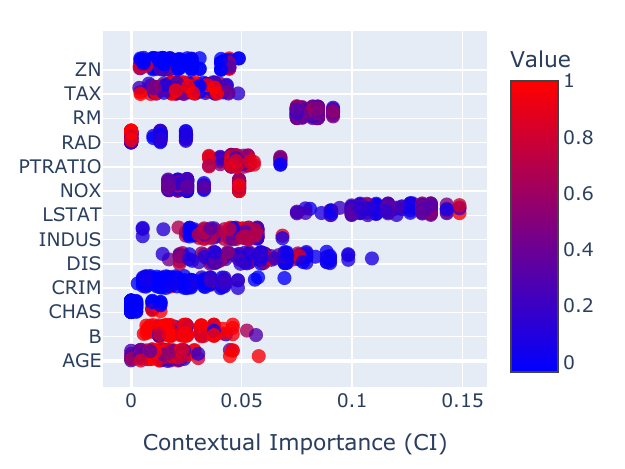} 
    \includegraphics[width=0.31\textwidth]{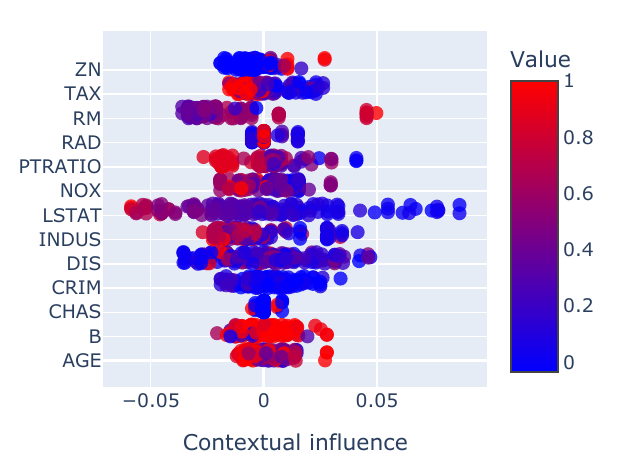} 
    \includegraphics[width=0.31\textwidth]{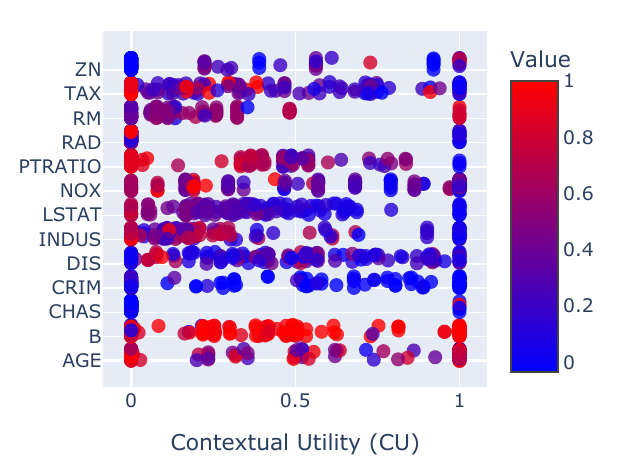}
    \caption{Beeswarm visualisations of CI, Contextual influence and CU for Boston Housing data set. }
    \label{fig:BostonBeeswarm}
\end{figure*}

Beeswarm plots give an overview of an entire data set by showing CI/CU/influence values of every feature and every instance. As in \url{https://github.com/slundberg/shap}\footnote{However, when visited on 27 April 2024, that page no longer uses Boston Housing but rather the California housing data set.}, we use the Boston data set and a Gradient Boosting model. The dot colors in Figure~\ref{fig:BostonBeeswarm} represent the feature value. The CI beeswarm in Figure~\ref{fig:BostonBeeswarm} reveals, for example, that the higher the value of \texttt{lstat} (\% lower status of the population), the more importance is given to the \texttt{lstat} feature as shown by higher CI values for high \texttt{lstat} values. The influence plot reveals that a high \texttt{lstat} value lowers the predicted home price, which is also shown in the CU plot where a high \texttt{lstat} value gives a low CU value. We use $CU_{ref}=0.390$ for Contextual influence, which corresponds to the average price so the reference value is the same as for the Shapley value. 

Unfortunately, we have not yet found a Python package that would produce beeswarm plots of similar quality as the SHAP package, as seen \eg at \url{https://github.com/shap/shap}. Therefore, we reproduce the beeswarm produced by the CIU \texttt{R} package in Figure~\ref{Fig:BeeswarmBostonR}, which is nearly identical to the one produced by SHAP. Similar beeswarm plots can also be produced for LIME and other XAI methods, which gives them some ``global''  explanation capabilities that are sometimes attributed only to SHAP. 

\subsection{Intermediate Concepts}

\begin{wrapfigure}{r}{0.6\columnwidth}
    \centering
    \includegraphics[width=0.6\columnwidth]{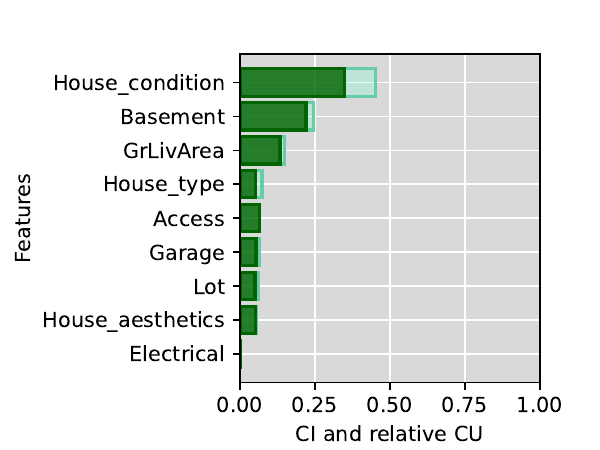} 
    \caption{Top-level PI plot for Ames housing data set.}
    \label{fig:Ames_CIU_top_level}
\end{wrapfigure}

CIU is not constrained to one input feature versus an output value. In reality, CIU can be used for any coalition of features $\{i\}$ relative to a coalition of features $\{I\}$, where $\{i\} \subseteq \{I\}$ and $\{I\} \subseteq \{1,\dots,n\}$. Furthermore, dependencies between features in $\{i\}$ and $\{I\}$ are automatically taken into account by how CIU is defined, as shown in \cite{FramlingAJCAI2022}. When we give names to such coalitions of features, then they can be used as Intermediate Concepts (IC) in explanations. By using ICs, we can build explanation vocabularies that correspond to the explainee's background and preferred levels of abstraction. The explanations and the explanatory interaction can also be adjusted accordingly. ICs could also be formed by studying feature dependencies; However, we believe it is more useful to focus on the explainees' needs. 





\begin{lstlisting}[style=mypystyle, caption={Example vocabulary for  Ames housing data set.}, label=lst:AmesVoc]
ames_voc = {
    "Garage":[c for c in df.columns if 'Garage' in c],
    "Basement":[c for c in df.columns if 'Bsmt' in c],
    "Lot":list(df.columns[[3,4,7,8,9,10,11]]),
    "Access":list(df.columns[[13,14]]),
    "House_type":list(df.columns[[1,15,16,21]]),
    "House_aesthetics":list(df.columns[[22,23,24,25,26]]),
    "House_condition":list(df.columns[[20,18,21,28,19,29]]),
    "First_floor_surface":list(df.columns[[43]]),
    "Above_ground_living area":[c for c in df.columns if 'GrLivArea' in c]
}
\end{lstlisting}

\begin{wrapfigure}{r}{0.6\columnwidth}
    \centering
    \includegraphics[width=0.6\columnwidth]{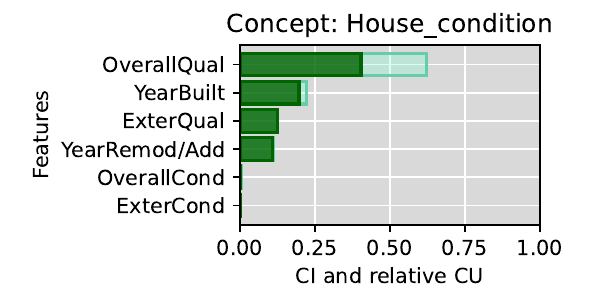} 
    \caption{PI plot for Intermediate Concept ``House condition''.}
    \label{fig:Ames_CIU_house_condition}
\end{wrapfigure}

Listing~\ref{lst:AmesVoc} shows a simple vocabulary defined for the Ames housing data set. Ames housing is a data set with 2930 houses described by 81 features. A gradient boosting model was trained to predict the sale price based on the 80 other features. With 80 features a ``classical'' bar plot explanation becomes unreadable. Furthermore, many features are strongly correlated, which causes misleading explanations because individual features have a small importance, whereas the joint importance can be significant. ICs solve these challenges, as illustrated in Figure~\ref{fig:Ames_CIU_top_level} that shows the top-level explanation for an expensive house. The corresponding explanation for the IC ``House condition'' is shown in Figure~\ref{fig:Ames_CIU_house_condition}. This vocabulary has been constructed based on common-sense knowledge about houses but the vocabulary to use could be adapted to individual explainees and updated during an explanation interaction. 

The plot in Figure~\ref{fig:Ames_CIU_top_level} is produced with the commands shown in Listing~\ref{lst:AmesTopCode}. The first line creates the CIU explainer and the second line sets to vocabulary to use. In the third line, we call \texttt{explain\_voc()} instead of \texttt{explain()} in order to use the vocabulary for the explanation. 
Figure~\ref{fig:Ames_CIU_house_condition} is produced with the commands shown in Listing~\ref{lst:AmesHouseConditionCode}. 

\begin{lstlisting}[style=mypystyle, caption={Producing a top-level explanation for Ames housing.}, label=lst:AmesTopCode]
CIU_ames = ciu.CIU(model.predict, ['Price'], data=X_train, out_minmaxs=out_minmaxs)
CIU_ames.vocabulary = ames_voc
top_res = CIU_ames.explain_voc(ames_instance, nsamples=10000)
CIU_ames.plot_ciu(top_res, plot_mode='overlap')
\end{lstlisting}

\begin{lstlisting}[style=mypystyle, caption={Producing an explanation for ``House condition'' Intermediate Concept.}, label=lst:AmesHouseConditionCode]
tconcept = "House_condition"
res = CIU_ames.explain(ames_instance, nsamples=1000, target_concept=tconcept, target_ciu=top_res)
CIU_ames.plot_ciu(res, plot_mode='overlap', figsize=(4,2), main=f"Concept: {tconcept}")
\end{lstlisting}

\subsection{Contrastive Explanations}\label{Sec:Contrastive}

Contrastive explanations compare one instance \texttt{A} with another instance \texttt{B}\footnote{For reasons of readability, we use \texttt{A} and \texttt{B} here for the two instances, rather than writing $x^{A}$ and $x^{B}$ or something similar.}, where \texttt{A} and/or \texttt{B} can be real or counterfactual. Humans often use and expect contrastive explanations that answer questions such as ``Why alternative A rather than B?'' or ``Why not alternative B rather than A?''. Just showing the explanation for \texttt{A} and the explanation for \texttt{B} is \textbf{not} a contrastive explanation. A contrastive explanation highlights the particular differences between \texttt{A} and \texttt{B}. In practice, this signifies that \texttt{A} is the reference against which we compare \texttt{B}. 

\begin{wrapfigure}{r}{0.7\columnwidth}
    \centering
    \includegraphics[width=0.7\columnwidth]{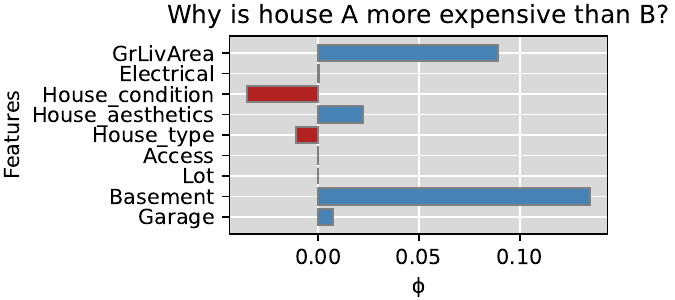} 
    \caption{Left: Contrastive ``Why?'' explanation for two expensive Ames houses. Right: Top-level counterfactual explanation for Ames instance \#1638.}
    \label{fig:Ames_CIU_contrastive}
\end{wrapfigure}

Since any value in the range $[0,1]$ can be used for $CU_{ref}$ in Equation~\ref{Eq:Contextual_influence}, CU values of instance \texttt{A} can also be used when producing a contextual influence explanation for \texttt{B}.  Figure~\ref{fig:Ames_CIU_contrastive} shows a contrastive explanation for why the previously used Ames instance with the predicted price of \$740 222 is more expensive than another house that has a predicted price of \$568 000. Contrastive values are in the range $[-1,1]$ by definition, so here again the actual $\phi$ values can be interpreted. For instance, the Basement of \texttt{A} is about 14\% better than the Basement of \texttt{B}. One might ask oneself what ``14\%'' signifies for a Basement but it is clearly better. Since Basement is an IC, it is also possible to obtain a detailed explanation for it. 

The contrastive explanation was produced by the code in Listing~\ref{lst:AmesContrastive}. In this contrastive explanation we used the same vocabulary as before. CIU results produced by \texttt{explain\_voc()} and the basic \texttt{explain()} methods are identical in format, so they can be passed to all plotting functions as such, including the \texttt{plot\_contrastive()} function.

%
\begin{lstlisting}[style=mypystyle, caption={Producing a textual explanation with Markdown style.}, label=lst:AmesContrastive]
inst1 = ames_instance
inst2 = X_test.iloc[[302]]
CIUres1 = CIU_ames.explain_voc(inst1)
CIUres2 = CIU_ames.explain_voc(inst2)
ciuplots.plot_contrastive(CIUres1, CIUres2, main="Why is house A more expensive than B?")
\end{lstlisting}


\section{Conclusion}\label{Sec:Conclusion}

The paper presents a Python implementation of the CIU method with the intention to provide researchers with an easy way to compare CIU with other XAI methods, and potentially even be used in commercial products. 
Furthermore, the paper shows capabilities that are specific to CIU and that presumably are not feasible or have not been exploited with currently popular methods such as LIME and SHAP. It is not possible to compare such CIU-specific capabilities against other methods, which is the reason for the lack of theoretical or empirical comparisons between CIU and other XAI methods. Regarding comparable capabilities such as those presented in Section~\ref{Sec:SotA}, comparisons have been made using the \texttt{R} version of CIU and presented \eg in \cite{FramlingAJCAI2022,Framling_xAI_2023}. 

Future work includes implementing CIU for use with natural language, time series and other kinds of input data. User studies regarding the understanding of different kinds of explanations by CIU and other XAI methods are also foreseen. Finally, our intention is to develop truly ``social XAI'' where the XAI system and the explainee could even engage in a co-constructive dialog \cite{Rohlfing2021} enabled by CIU's capability of generating explanations that answer different questions and by using ICs.

\bibliographystyle{named}
\bibliography{FramlingXAI_IJCAI2024}

\begin{thebibliography}{}

\bibitem[\protect\citeauthoryear{Anjomshoae \bgroup \em et al.\egroup }{2019}]{AnjomshoaeEtAl_CIU_2019}
Sule Anjomshoae, Kary Fr{\"a}mling, and Amro Najjar.
\newblock Explanations of black-box model predictions by contextual importance and utility.
\newblock In {\em Proc. $1^{st}$ international workshop on Explainable, transparent autonomous agents and multi-agent systems (EXTRAAMAS), Montreal, QC, Canada, May 13–14, 2019}, pages 95--109. Springer, 2019.

\bibitem[\protect\citeauthoryear{Anjomshoae \bgroup \em et al.\egroup }{2020}]{AnjomshoaeEtAl_PyCIU}
Sule Anjomshoae, Timotheus Kampik, and Kary Fr{\"a}mling.
\newblock {Py-CIU: A Python Library for Explaining Machine Learning Predictions Using Contextual Importance and Utility}.
\newblock In {\em {Proceedings of IJCAI-PRICAI 2020 Workshop on Explainable Artificial Intelligence (XAI)}}, 2020.

\bibitem[\protect\citeauthoryear{Fr{\"a}mling \bgroup \em et al.\egroup }{2021}]{Framling_ciuimageR_EXTRAAMAS_2021}
Kary Fr{\"a}mling, Samanta Knapi{\u c}, and Avleen Malhi.
\newblock {ciu.image: An R Package for Explaining Image Classification with Contextual Importance and Utility}.
\newblock In Davide Calvaresi, Amro Najjar, Michael Winikoff, and Kary Fr{\"a}mling, editors, {\em Explainable and Transparent AI and Multi-Agent Systems - 3rd International Workshop, EXTRAAMAS 2021, Revised Selected Papers}, Lecture Notes in Computer Science (including subseries Lecture Notes in Artificial Intelligence and Lecture Notes in Bioinformatics), pages 55--62, Germany, 2021. Springer.

\bibitem[\protect\citeauthoryear{Fr{\"a}mling \bgroup \em et al.\egroup }{2024}]{Framling_py_ciu_image_EXTRAAMAS_2024}
Kary Fr{\"a}mling, Ioan-Vlad Apopei, Gustav~Grund Pihlgren, and Avleen Malhi.
\newblock {py\textunderscore ciu\textunderscore image: a Python library for Explaining Image Classification with Contextual Importance and Utility}.
\newblock In {\em To appear: Explainable and Transparent AI and Multi-Agent Systems - 6th International Workshop, EXTRAAMAS 2024, Revised Selected Papers}, Germany, 2024. Springer.

\bibitem[\protect\citeauthoryear{Fr{\"a}mling}{1992}]{FramlingDEA_1992}
Kary Fr{\"a}mling.
\newblock {\em Les r{\'e}seaux de neurones comme outils d'aide {\`a} la d{\'e}cision floue}.
\newblock {D.E.A. thesis}, {INSA de Lyon}, July 1992.

\bibitem[\protect\citeauthoryear{Fr{\"a}mling}{1996a}]{FramlingAISB_1996}
Kary Fr{\"a}mling.
\newblock Explaining results of neural networks by contextual importance and utility.
\newblock In Robert Andrews and Joachim Diederich, editors, {\em Rules and networks: Proc. of Rule Extraction from Trained Artificial Neural Networks Workshop, AISB'96 conference}, Brighton, UK, 1-2 April 1996.

\bibitem[\protect\citeauthoryear{Fr{\"a}mling}{1996b}]{FramlingThesis_1996}
Kary Fr{\"a}mling.
\newblock {\em {Mod{\'e}lisation et apprentissage des pr{\'e}f{\'e}rences par r{\'e}seaux de neurones pour l'aide {\`a} la d{\'e}cision multicrit{\`e}re}}.
\newblock Phd thesis, {INSA de Lyon}, March 1996.

\bibitem[\protect\citeauthoryear{Fr{\"a}mling}{2021}]{Framling_XAI_WS_AAAI_2021}
Kary Fr{\"a}mling.
\newblock {Contextual Importance and Utility in R: the ‘ciu’ Package}.
\newblock In P.~Madumal, S.~Tulli, R.~Weber, and D.~Aha, editors, {\em Proc. $1^{st}$ Workshop on Explainable Agency in Artificial Intelligence Workshop, $35^{th}$ AAAI conference on Artificial Intelligence}, pages 110--114, 2021.

\bibitem[\protect\citeauthoryear{Fr{\"a}mling}{2022}]{FramlingAJCAI2022}
Kary Fr{\"a}mling.
\newblock {Contextual Importance and Utility: A Theoretical Foundation}.
\newblock In G.~Long, X.~Yu, and S.~Wang, editors, {\em AI 2022: Proc. $34^{th}$ Australasian Joint Conference on Artificial Intelligence}, pages 117--128. Springer, 2022.

\bibitem[\protect\citeauthoryear{Fr\"{a}mling}{2023}]{Framling_CIU_EXTRAAMAS_2023}
Kary Fr\"{a}mling.
\newblock {Counterfactual, Contrastive, and Hierarchical Explanations with Contextual Importance and Utility}.
\newblock In {\em {Explainable and Transparent AI and Multi-Agent Systems: 5th International Workshop, EXTRAAMAS 2023, London, UK, May 29, 2023, Revised Selected Papers}}, page 180–184, Berlin, Heidelberg, 2023. Springer-Verlag.

\bibitem[\protect\citeauthoryear{Friedman}{2001}]{Friedman_PDP_2001}
Jerome~H. Friedman.
\newblock {Greedy function approximation: A gradient boosting machine.}
\newblock {\em The Annals of Statistics}, 29(5):1189 -- 1232, 2001.

\bibitem[\protect\citeauthoryear{Främling}{2023}]{Framling_xAI_2023}
Kary Främling.
\newblock {Feature Importance versus Feature Influence and What It Signifies for Explainable AI}.
\newblock In Luca Longo, editor, {\em {Explainable Artificial Intelligence}}, pages 241--259, Cham, 2023. {Springer Nature Switzerland}.

\bibitem[\protect\citeauthoryear{Keeney and Raiffa}{1976}]{KeeneyRaiffa_1976}
Ralph Keeney and Howard Raiffa.
\newblock {\em Decisions with Multiple Objectives: Preferences and Value Trade-Offs}.
\newblock Cambridge University Press., 1976.

\bibitem[\protect\citeauthoryear{K{\"o}ksalan \bgroup \em et al.\egroup }{2011}]{koksalan2011multiple}
M.M. K{\"o}ksalan, J.~Wallenius, and S.~Zionts.
\newblock {\em Multiple Criteria Decision Making: From Early History to the 21st Century}.
\newblock World Scientific, 2011.

\bibitem[\protect\citeauthoryear{Lundberg and Lee}{2017}]{NIPS2017_Lundberg_XAI}
Scott~M Lundberg and Su-In Lee.
\newblock A unified approach to interpreting model predictions.
\newblock In I.~Guyon, U.~V. Luxburg, S.~Bengio, H.~Wallach, R.~Fergus, S.~Vishwanathan, and R.~Garnett, editors, {\em Advances in Neural Information Processing Systems 30}, pages 4765--4774. Curran Associates, Inc., 2017.

\bibitem[\protect\citeauthoryear{Lundberg \bgroup \em et al.\egroup }{2018}]{LundbergEtAl_TreeSHAP_2018}
Scott~M. Lundberg, Gabriel~G. Erion, and Su-In Lee.
\newblock Consistent individualized feature attribution for tree ensembles.
\newblock {\em CoRR}, abs/1802.03888, 2018.

\bibitem[\protect\citeauthoryear{Ribeiro \bgroup \em et al.\egroup }{2016}]{ribeiro2016_LIME}
Marco~Tulio Ribeiro, Sameer Singh, and Carlos Guestrin.
\newblock {"Why Should I Trust You?": Explaining the Predictions of Any Classifier}.
\newblock In {\em Proceedings of the 22nd ACM SIGKDD International Conference on Knowledge Discovery and Data Mining}, KDD '16, page 1135–1144, New York, NY, USA, 2016. {Association for Computing Machinery}.

\bibitem[\protect\citeauthoryear{Rohlfing \bgroup \em et al.\egroup }{2021}]{Rohlfing2021}
Katharina~J. Rohlfing, Philipp Cimiano, Ingrid Scharlau, Tobias Matzner, Heike~M. Buhl, Hendrik Buschmeier, Elena Esposito, Angela Grimminger, Barbara Hammer, Reinhold Häb-Umbach, Ilona Horwath, Eyke Hüllermeier, Friederike Kern, Stefan Kopp, Kirsten Thommes, Axel-Cyrille Ngonga~Ngomo, Carsten Schulte, Henning Wachsmuth, Petra Wagner, and Britta Wrede.
\newblock Explanation as a social practice: Toward a conceptual framework for the social design of ai systems.
\newblock {\em IEEE Transactions on Cognitive and Developmental Systems}, 13(3):717--728, 2021.

\bibitem[\protect\citeauthoryear{Roy}{1985}]{roy1book85}
B.~Roy.
\newblock {\em M\'ethodologie multicrit\`ere d'aide \`a la d\'ecision}.
\newblock Economica, Paris, 1985.

\bibitem[\protect\citeauthoryear{Vincke}{1992}]{Vin92}
{Ph}. Vincke.
\newblock {\em Multicriteria Decision-Aid}.
\newblock J. Wiley, New York, 1992.

\end{thebibliography}

\end{document}